\documentclass[letterpaper, 10 pt, conference]{ieeeconf}% % \usepackage{generic}
\IEEEoverridecommandlockouts   
\usepackage{bm}        % for math
\usepackage[backend=biber,sorting=none]{biblatex}
\usepackage[english]{babel}

\usepackage{quiver}
\usepackage[most]{tcolorbox}
\usepackage{xcolor}
\colorlet{shadecolor}{orange!15}
\usepackage{stmaryrd}
\usepackage{mathtools}
\usepackage{algorithm} 
\usepackage{algpseudocode} 

\usepackage{authblk}
\usepackage{float}
\usepackage{latexsym,amsmath,amscd,amssymb,graphics,wasysym}
\usepackage{enumerate}
\usepackage{wrapfig}
\usepackage{graphicx}
\usepackage{framed}
\usepackage{url}
\usepackage{color}
\usepackage{cancel}
\usepackage[utf8]{inputenc} %useful to type directly diacritic characters
\makeatletter
\def\widebreve#1{\mathop{\vbox{\m@th\ialign{##\crcr\noalign{\kern3\p@}%
      \brevefill\crcr\noalign{\kern3\p@\nointerlineskip}%
      $\hfil\displaystyle{#1}\hfil$\crcr}}}\limits}

\def\brevefill{$\m@th \setbox\z@\hbox{$\braceld$}%
  \bracelu\leaders\vrule \@height\ht\z@ \@depth\z@\hfill\braceru$}
\makeatletter

\newcommand{\tl}[1]{\widetilde{#1}}

\newcommand{\pp}[2]{\frac{\partial #1}{\partial #2}}

\newcommand{\dd}[2]{\frac{d #1}{d #2}}

\newcommand{\bs}{\boldsymbol{s}}
\newcommand{\bepsilon}{\boldsymbol{\epsilon}}

\newcommand{\bO}{\boldsymbol{O}}

\newcommand{\bA}{\boldsymbol{A}}

\newcommand{\ba}{\boldsymbol{a}}
\newcommand{\balpha}{\boldsymbol{\alpha}}
\newcommand{\bSig}{\boldsymbol{\Sigma}}

\newcommand{\bZ}{\boldsymbol{Z}}

\newcommand{\bW}{\boldsymbol{W}}

\newcommand{\bmu}{\boldsymbol{\mu}}

\newcommand{\bpsi}{\boldsymbol{\psi}}

\newcommand{\bS}{\boldsymbol{S}}
\newcommand{\bphi}{\boldsymbol{\varphi}}

\newcommand{\bF}{\boldsymbol{F}}

\newcommand{\bH}{\boldsymbol{H}}

\newcommand{\btheta}{\boldsymbol{\theta}}
\newcommand{\bX}{\boldsymbol{X}}

\newcommand{\bzero}{\boldsymbol{0}}

\newcommand{\bI}{\mathbb{I}}
\newcommand{\bh}{\boldsymbol{h}}

\newcommand{\om}{\omega}

\newcommand{\mbG}{\mathbb{G}}

\newcommand{\mbQ}{\mathbb{Q}}

\newcommand{\mbR}{\mathbb{R}}

\newcommand{\mbE}{\mathbb{E}}

\newcommand{\mcN}{\mathcal{N}}
\newcommand{\mcS}{\mathcal{S}}

\newcommand{\mcA}{\mathcal{A}}

\newcommand{\mcF}{\mathcal{F}}

\newcommand{\mcL}{\mathcal{L}}

\newcommand{\mcT}{\mathcal{T}}

\newcommand{\eps}{\varepsilon}

\newtheorem{remark}{Remark}

\newtheorem{problem}{Problem}

\newcommand{\rem}[1]{}

\usepackage[utf8]{inputenc}

\title{\LARGE \bf Learning to Crawl: Latent Model-Based Reinforcement Learning for Soft Robotic Adaptive Locomotion}
\author{Vaughn Gzenda$^1$ and Robin Chhabra$^1$ 
\thanks{*This work was supported by a grant from the Natural Sciences and Engineering Research Council of Canada (DGECR-2019-00085).}
\thanks{$^1$Embodied Learning and Intelligence for eXploration and Innovative soft Robotics (ELIXIR) Lab, Toronto Metropolitan University, Toronto, ON, Canada. 
{\tt\small 	vaughn.gzenda@torontomu.ca}, {\tt\small robin.chhabra@torontomu.ca}}
}

\begin{document}

\maketitle

\begin{abstract}
Soft robotic crawlers are mobile robots that utilize soft body deformability and compliance to achieve locomotion through surface contact. Designing control strategies for such systems is challenging due to model inaccuracies, sensor noise, and the need to discover locomotor gaits. In this work,
we present a model-based reinforcement learning (MB-RL) framework in which
latent dynamics inferred from onboard sensors serve as a predictive model that
guides an actor-critic algorithm to optimize locomotor policies. We evaluate
the framework on a minimal crawler model in simulation using inertial measurement units and time-of-flight sensors as observations. The learned latent
dynamics enable short-horizon motion prediction while the actor-critic discovers
effective locomotor policies. This approach highlights the potential of latent-dynamics MB-RL for enabling embodied soft robotic adaptive locomotion based solely on noisy
sensor feedback.
    % % Topic
    % Soft robotic crawlers (SRCs) are mobile robots that utilize soft body deformability  and compliance to achieve locomotion through surface contact. 
    % % motivation
    % Designing control strategies for such systems is challenging due to model inaccuracies, sensor noise, and the need to generate effective locomotor gaits. 
    % % contribution 
    % In this work, we propose a model-based reinforcement learning (MB-RL) framework for SRC locomotion in which latent dynamics are inferred directly from onboard sensors and are used to optimize gait policies for forward locomotion.
    % % detail/nuance
    % Our approach extends existing MB-RL architectures, such as Dreamer, by representing the translational kinematics of the rigid-body motion of the entire robot into the latent state and using these dynamics for trajectory forecasting and gait parameter learning using an actor-critic method. 
    % % evidence/contribution
    % We evaluate the framework on a minimal crawler model in simulation using inertial measurement units and time of flight sensors as observations. 
    % % narrow impact
    % The learned latent dynamics enable motion prediction over a horizon and the actor-critic methodology enables the discovery of effective gaits.
    % % broad impact 
    % More broadly, this work highlights the potential of latent dynamics MB-RL for enabling soft robotic locomotion based solely on noisy sensor feedback. 
\end{abstract}

\section{Introduction}

% motivation: locomotion for some robotic systems

% inchworm crawler locomotion
% Active Inference, control as inference
% ch

% Introduction 
Soft Robotic Crawlers (SRCs) \cite{asawalertsak2023frictional,chang2021shape,joey2017earthworm,shen2024optimal} represent a class of mobile robots that exploit the inherent compliance and deformability of soft materials to achieve robust, adaptive locomotion in complex and unstructured environments. Unlike traditional rigid-bodied robots, soft crawlers can conform to terrain, absorb impacts, and navigate through confined spaces, making them particularly well-suited for applications in search and rescue, environmental/industrial monitoring, biomedical devices, and exploration in hazardous or delicate settings.

% Soft robots

% Soft Robotic Crawler Design 
Biological inspiration often plays a central role in the design of soft crawling systems, with organisms such as worms, caterpillars, and snakes providing models. Common actuators include the expansion and contraction of fluidic/pneumatic chambers, electromagnetic material properties, and Shape Memory Alloys (SMAs) \cite{chang2021shape,pan2025bioinspired}. In one design \cite{joey2017earthworm}, an earthworm inspired crawler actively modulates its contact friction forces to anchor the base and/or head in place and internal shape is extended/retracted to achieve forward locomotion. Another design includes two linear pneumatic actuators able to also perform turning in addition to modulating contact forces \cite{even2023locomotion}. Some soft crawler robots design the contact substrate to have two friction coefficients to achieve locomotion by only modulating the shape of the body making use of anisotropic friction \cite{asawalertsak2023frictional,chang2021shape}. A multi-modular design of an inchworm with suction cups at the contact points is capable of both horizontal and vertical locomotion \cite{zhang2021inchworm}. Soft robotic crawlers have been proposed for applications in industrial monitoring such as pipe inspection \cite{lin2023single,zhang2021inchworm,zhang2019design}. A review of the design of soft robots for locomotion can be found in \cite{sun2021soft}.

% Soft robot gait planning
Gait planning in soft robotics is a central challenge due to the interplay between the compliant body mechanics, environmental interactions, and often high dimensional control spaces.
A classical approach for inchworm robot gait planning is to use a finite state space representation of the anchoring points and extended/contracted body states \cite{chen2001locomotive}. Another study of soft crawler locomotion uses a simplified planar rigid-link models with elastic joints to plan the gait sequencing \cite{gamus2020understanding}. 
This work \cite{chang2021shape} develops a shape-based modeling framework that enables predictive gait planning and control for an SMA-actuated inchworm soft robot. Gait planning for a soft multi-legged robot has also been studied and implemented on starfish like robot \cite{shepherd2011multigait}.

% soft robotic control
A main difficulty in soft robot control arises from balancing accurate modeling of body deformations with the practical requirements of real-time control. Finite element approaches \cite{ferrentino2023finite} offer more accuracy in a tradeoff with real-time prediction. Finite difference approaches \cite{jilani2024solving} are better suited for real-time prediction, but are numerically unstable under contact forces. In the middle ground, simplified models such as mass-spring-damper models \cite{shen2024optimal} or lumped mass models \cite{zhang2024adaptive} attempt to balance accuracy with simplicity for control design. Using a minimal model of a SRC, a hill-climbing algorithm is developed to solve an optimal periodic control problem to optimize the gait of the robot \cite{shen2024optimal}. A perturbation approach \cite{gusty2025optimal}, with a Bernoulli beam model of the robot and nonlinear friction, splits the dynamics into two time scales and performs open loop control of a gait. A model based control of a worm robot actuating its body for peristaltic locomotion has been developed for pipe inspection\cite{riddle20253d}. This work demonstrates how neural network gradients can be used to derive a linearized state-space model for soft robots, enabling effective model predictive control \cite{gillespie2018learning}. Central Pattern Generator (CPG)-based control to regulate gait frequency in a SRC, enabling adaptable locomotion and simple speed regulation \cite{asawalertsak2023small}.

% Model based RL
Model Predictive Control (MPC) \cite{garcia1989model} is widely regarded as a standard tool in robotic motion planning and control. For soft robots, however, the reliance on accurate and tractable models often limits its applicability, spurring research into data-driven alternatives that learn system dynamics from observations. Probabilistic models such as PILCO \cite{deisenroth2011pilco} offer data efficiency but scale poorly to high-dimensional systems, while Probabilistic Ensembles with Trajectory sampling (PETs) leverage neural network ensembles and rollouts within MPC to produce robust plans \cite{chua2018deep}.

% Latent Dynamics models
Latent dynamics models in Model-Based Reinforcement Learning (MB-RL) have been developed to enable planning directly from high-dimensional observations such as pixels, with notable examples including PlaNet \cite{hafner2019learning}, Dreamer \cite{hafner2019dream}, and extensions incorporating discrete latent states and improved planning strategies \cite{hafner2020mastering, hafner2023mastering}. More broadly, deep learning and reinforcement learning approaches have advanced the development of world models for decision making \cite{matsuo2022deep}, with recent works such as robotic world models \cite{li2025robotic} and DayDreamer \cite{wu2023daydreamer} demonstrating their ability to support long-horizon planning and physical robot learning. Building on these advances, latent dynamics and world models provide a promising foundation for soft robotic control, where learning compact representations of complex dynamics can facilitate gait planning and adaptive locomotion in soft robotic crawlers.

% Contributions 
In this paper, we present a latent dynamics modeling framework for soft robotic crawlers. The model is learned directly from noisy sensor streams, specifically Inertial Measurement Units (IMUs) and Time-Of-Flight (TOF) sensors, without relying on explicit analytical models of the robot’s complex continuum body. By capturing the underlying latent dynamics, the model enables forecasting of the crawler’s future behavior under different control inputs. 
This predictive capability is then leveraged to train an actor–critic reinforcement learning framework, which optimizes the robot’s locomotion strategy. In particular, the learned latent model serves as a surrogate for real-world interaction, allowing the policy to efficiently explore candidate gaits and converge toward those that maximize forward displacement.  
The main contributions of this work are summarized as follows:
\begin{itemize}
    \item We develop a data-driven latent dynamics model for soft robotic crawlers that is estimated directly from noisy IMU and TOF measurements.
    \item We integrate the learned latent model with an actor–critic reinforcement learning framework to discover gaits that maximize forward locomotion.
\end{itemize}

This paper is structured as follows. In Section \ref{Crawler_system} review the dynamics of soft robotic crawlers and introduce the IMU and TOF sensors. In Section \ref{MB_RL} we describe the latent model based reinforcement learning framework for gait generation. In Section \ref{Simulation_studies} we perform simulation studies. Lastly, in Section \ref{Conclusion} we end with some concluding remarks.  
% \begin{itemize}
%     \item Developing a latent state transition model for soft crawler dynamics estimated from sensor measurements and using this model for gait planning. 
% \end{itemize}

% In this paper, we study a latent dynamics model approach for soft robotic crawler locomotion using noisy inertial measurement unit and time of flight sensor observations. 

% Biological inspiration often plays a central role in the design of soft crawling systems, with organisms such as worms, caterpillars, and octopi providing models for distributed actuation, decentralized control, and mechanically intelligent body structures. These systems rely on diverse locomotion strategies—ranging from peristaltic waves and shape-morphing gaits to friction modulation and body anchoring—to generate propulsion and maneuverability.

% Despite significant progress in materials science, fabrication techniques, and control architectures, the development of efficient, controllable, and scalable soft crawling robots remains an open challenge. Key research questions include how to model the complex nonlinear dynamics of soft structures, how to design actuation and control strategies that leverage morphological computation, and how to integrate sensing and feedback for adaptive behavior in uncertain environments.

% ==============================
% Preliminaries
% ==============================
\section{Crawler Robotic System}\label{Crawler_system}
% ==============================
% ==============================
% ================================================================================

% ==============================
% ==============================
% ================================================================================
\subsection{One Dimensional Crawler Dynamics}

We consider a minimal model of soft robotic locomotion as a one dimensional inchworm robot such as in \cite{das2023earthworm,joey2017earthworm}. The configuration space for the crawler is $Q = \mbR\times \mbR$ with local coordinates $(x_1,x_2)$ for the location of the base and head, respectively. We define the velocity of the base and head by $(v_1,v_2)$. Let $m_1$ be the mass of the base and let $m_2$ be the mass of the head, and let elasticity constant and damping $b$ for joint connecting the base to the head. Let $F_u = (B_u\,u,-B_u\,u)$ be the control inputs. The contact forces are treated as signed anisotropic friction $F_f =(F_{\sigma1}\,\sigma(v_1),F_{\sigma2}\sigma(v_2))$ modelled by sigmoid function \cite{shen2024optimal} 
\begin{align}
    \sigma(v_i) =\frac{1}{2}\left(\frac{1 + N_f}{1 + e^{-(-v_i - v_{\text{offset},i})/\eps_f}} + 1-N_f \right)
\end{align} where $(N_f,\eps_f)$ are tuning parameters for the anisotropic friction and $v_{\text{offset},i}$ is defined such that $\sigma(0) = 0$ and $F_{\sigma,i}$ are the amplitude of the friction forces.

\begin{figure}[H]
    \centering
    \includegraphics[width=0.75\linewidth]{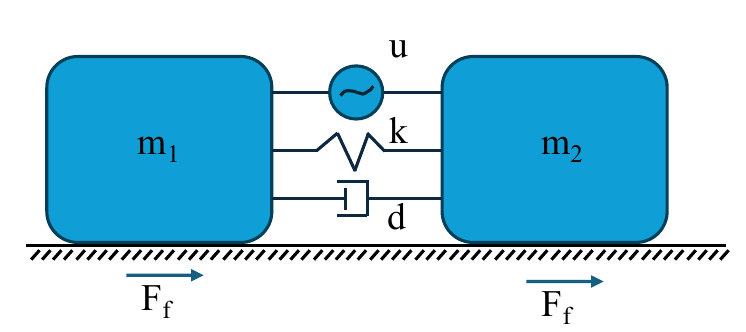}
    \caption{Model of Soft Robotic Crawler}
    \label{fig:placeholder}
\end{figure}

The dynamics of the one dimensional crawler is given by 
\begin{align} \label{generic_crawler_dynamics}
    \dd{}{t}\begin{bmatrix}
        x_1 \\ x_2 \\ m_1v_1 \\ m_2v_2 
    \end{bmatrix}\! = \!\!\begin{bmatrix}
        v_1 \\ v_2 \\ -k(x_1 - x_2) - b(v_1-v_2) + F_{\sigma1}\sigma(v_1) + B_u\,u\\
        k(x_1 - x_2) + b(v_1-v_2) + F_{\sigma2}\sigma(v_2) - B_u\,u
    \end{bmatrix}
\end{align}

\subsection{Internal/Rigid Decomposition} The one dimensional motion of the robot can be decomposed into the rigid motion of the centre of mass $s_1$ and the internal strain of the robot $s_2$ defined by 
\begin{align} 
    s_1 = \frac{1}{M}(m_1x_1 + m_2x_2), \quad s_2 = x_2 - x_1.
\end{align} where $M = m_1 + m_2$ is the total mass. Further, we define the velocities of the centre of mass by $w_1$, and strain velocity $w_2$ by 
\begin{align} \label{mechanical_change_cords}
    & w_1 = \frac{1}{M}(m_1v_1 + m_2v_2), \quad w_2 = v_2 - v_1.
\end{align} Under this change of coordinates, we find the dynamics of the crawler can be decomposed into rigid and internal motion, where the rigid motion is given by 
\begin{align}
    \dd{}{t}\begin{bmatrix}
        s_1 \\ w_1 
    \end{bmatrix} & = \begin{bmatrix}
        w_1 \\ \frac{1}{M}\left(F_{\sigma1}\Sigma_1(w_1,w_2) + F_{\sigma2}\Sigma_2(w_1,w_2) \right) 
    \end{bmatrix}\\ \notag
    & =: \bF_1(w_1,w_2)
\end{align} where $M = m_1 +m_2$ is the total mass and 
\begin{align}
    &\Sigma_1(w_1,w_2) = \sigma\left(\frac{1}{M}(M w_1 - m_2 w_2) \right)\\
    &\Sigma_2(w_1,w_2) = \sigma\left(\frac{1}{M}(M w_1 + m_1 w_2) \right)
\end{align} are the friction forces in centre of mass/strain coordinates. The internal motion of the strain of the robot is given by 
\begin{align}
    \dd{}{t}\begin{bmatrix}
        s_2 \\ w_2
    \end{bmatrix} &= \begin{bmatrix}
        w_2 \\ -k\mu s_2 -b\mu w_2 + \frac{F_{\sigma1}}{m_2}\Sigma_1 -\frac{F_{\sigma2}}{m_1}\Sigma_2  + B_u\mu\,u 
    \end{bmatrix}
\\ \notag
&=\bF_2(s_2,w_1,w_2,u)
\end{align} where $\mu= 1/m_1 + 1/m_2$.
\begin{remark}
    The rigid/internal decomposition is obtained through a change of coordinates, as defined in~\eqref{mechanical_change_cords}, which arises from the mechanical connection~\cite{marsden1992lectures}, given by 
    \begin{align}
        \mcA = \frac{1}{M}\begin{bmatrix}
            m_1 & m_2
        \end{bmatrix}.
    \end{align} 
    The existence of the mechanical connection is a result of the translational invariance of the (crawler) locomotion system \cite{marsden1992lectures,marsden2006hamiltonian,marsden2013introduction}. 
    % The mechanical connection gives a change of coordinates 
    
    % Any locomotion system  As the motion is one dimensional motion along a line, it can easily shown that the connection is flat.
\end{remark}

The uncertainty in the motion of the true crawler system is described by Stochastic Differential Equations (SDEs) that are modified from the deterministic dynamics
\begin{align}
    \mbox{d}\begin{bmatrix}
        s_1 \\w_1 \\ s_2 \\ w_2 
    \end{bmatrix} = \begin{bmatrix}
        \bF_1(w_1,w_2)\\ \bF_2(s_2,w_1,w_2,u)
    \end{bmatrix}\mbox{d}t + \mbG\circ\mbox{d}\bW_t
\end{align} where $\mbG$ is the noise shaping matrix and $\bW_t$ is a Wiener process. The Fokker-Planck equations associated with the SDEs have the evolution of the mean and variance given by the following ordinary differential equations 
\begin{align}
    &\dd{}{t} \begin{bmatrix}
        \mu_{s_1} \\ \mu_{w_1} \\ \mu_{s_2} \\ \mu_{w_2}
    \end{bmatrix} = \begin{bmatrix}
        \bF_{1}(\mu_{w_1},\mu_{w_2})\\ \bF_{2}(\mu_{s_2},\mu_{w_1},\mu_{w_2},u)
    \end{bmatrix}\\
    &\dd{}{t}\bSig_{s,w} = \pp{\bF}{\bS}^T\bSig_{s,w} + \bSig_{s,w}\pp{\bF}{\bS}
\end{align} where $\mcS = (s_1,w_1,s_2,w_2)$ are the state variables. Given a sampling time $\Delta t$, the evolution of the mean and covariance may be discretized $\tl{\bmu}_{s,w},\,\tl{\bSig}_{s,w}$ using an integrator such as Euler or RK4 
\begin{align}
    & \tl{\bmu}_{s,w}(t+\Delta t) = \tl{\bmu}_{s,w}(t) + \bF_{\mu}(\tl{\bmu}_{s,w}(t); \Delta t)\\
    & \tl{\bSig}_{s,w}(t+\Delta t) = \tl{\bSig}_{s,w}(t) + \bF_{\Sigma}(\tl{\bmu}_{s,w}(t); \Delta t)
\end{align} providing a Gaussian model of the evolution of the state transition probabilities 
\begin{align}
    p(\mcS_{t+1}| \mcS_{t},u_{t}) \sim \mcN(\tl{\bmu}_{s,w,t},\tl{\bSig}_{s,w,t}).
\end{align} 

\subsection{Sensor Models}
\paragraph{Inertial Sensor} For the crawler's sensor of its motion, we employ an Inertial Measurement Unit (IMU) placed on both the head and the base of the crawler described by  
\begin{align}
\quad \alpha_i(t) = \bar{a}_i(t) + b_{a_i}(t) + \eta_{i}(t) 
\end{align} 
where $\alpha_i(t)$ is the measured IMU reading, $\bar{a}_i(t)$ is the true acceleration, $b_{a_i}(t)$ is the bias, and $\eta_i(t)$ is zero mean gaussian noise.

\paragraph{Proprioceptive Sensors} For the crawler to have sensor of its location, we place a TOF sensor on the base and the head of the crawler. Let $X_{i,0}$ be the location of an object for the TOF sensor to measure against, then a model for the sensor reading $Z_i$ is 
\begin{align}
    X_i = |X_{i,0} - x_i| + \eta_{TOF}.
\end{align} where $\eta_{TOF}$ is zero mean Gaussian noise, we collect the TOF observations in a vector $\bX_t = (X_{1}(t),X_{2}(t))$. 

% The time of flight sensor $\by_t = (X_1,X_2)_t$ has the state measurement probabilities 
% \begin{align}
%     p(\by_{t}| \bx_t,\bv_t) \sim \mcN( h(\bx_t), \bSig_{\by}).
% \end{align}

% By filtering the TOF sensor we estimate the average velocity over a sample time $\delta t$ to be 
% \begin{align}
%     V_i = \frac{X_{i}(t + \delta t) - X_i(t)}{\delta t}.
% \end{align} 

% We define a distinguished light sensor to represent the goal/reward for motion. 
% \begin{align}
%     L_i = \frac{\alpha_0}{|x_i -L_{i,0}|^2 + \eps}
% \end{align} The crawler is designed to be drawn to the light sensor. The stochastic equations describing the light measurement model is given by 
% \begin{align} \label{Light_sensor_SDE}
%      L_i = \frac{\alpha_0}{|x_i -L_{i,0}|^2 + \eps} + \eta_{L}.
% \end{align}

% ===============
\section{Model-Based Reinforcement Learning for Crawler Locomotion}\label{MB_RL}

We assume that we do not have access to the crawler dynamics and instead only have access to the IMU and TOF measurements in addition to the control inputs. Using these sensors and insights into the state space structure of the crawler dynamics discussed in the previous section, we learn a latent dynamics model by minimizing the variational free energy (negative evidence lower bound) of a model conditioned on the control action with a model conditioned on the sensor observations. 

\subsection{Perception Module} For Bayesian Inference of the state transition motion, we seek to estimate a posterior probability distribution $p(X_t|D_t)$ may be estimated from observed data $D_t$ by Bayes rule
\begin{align}
    p(X_t|D_t) =\frac{p(D_t,X_t)}{p(D_t)} = \frac{p(D_t|X_t)p(X_t)}{\int p(D_t,X)dX}.
\end{align} However, this integral is intractable to solve, so we approximate the posterior distribution by another distribution $q(X_t)$ and minimize the associated free energy defined by the functional
\begin{align}
    \mcF_t = D_{KL}[q(X_t)||p(X_t|D_t)] - \log \, p(D_t).
\end{align} where $D_{KL}$ is the Kullback–Leibler divergence. We use the variational free energy to learn a state transition model $p(\mcS_{t+1}| \mcS_{t},u_{t})$ from sensor measurements. 

\begin{remark}
    As the Kullback–Leibler is strictly non-negative, the variational free energy is thus an upper bound on the surprise $-\log p(D_{t})$ from the observed data $D_t$. Therefore, the agent seeks to minimize its surprise by minimizing the variational free energy. 
\end{remark}

The state transition depends on both the control inputs as well as the observations. However, a priori we do not know the specific functional form of the state dynamics, hence we represent the dynamics of the robot with a parameterized vector field $\bF_{\varphi} = (\bF_{1,\varphi},\bF_{2,\varphi})$ given in the form 
\begin{align}
    \mbox{d}\begin{bmatrix}
        s_1 \\ w_1 \\ s_2 \\ w_2
    \end{bmatrix} = \begin{bmatrix}
        \bF_{1}(\bs,\boldsymbol{w},\bphi)\\
        \bF_{2}(\bs,\boldsymbol{w},u,\bphi)
    \end{bmatrix}dt + \mbQ\circ \mbox{d}\bW
\end{align}  The Fokker-Plank equations associated with these SDEs has zeroth order approximation of the mean and covariance given by 
\begin{align} \label{eom_paremeterized:eq1}
    & \dd{}{t}\bmu_{\mcS} = \begin{bmatrix}
        \bF_1(\bmu_{\mcS},\bphi_{1})\\ \bF_2(\bmu_{\mcS},u,\bphi_{2})
    \end{bmatrix}\\ \label{eom_paremeterized:eq2}
    &\dd{}{t}\bSig_{s,\nu} = \pp{\bF}{\bmu_{\mcS}}^T\bSig_{s,\nu} + \bSig_{s,\nu}\pp{\bF}{\bmu_{\mcS}}
\end{align} where $\bmu_{\mcS} = (\mu_{s_1},\mu_{w_1},\mu_{s_2},\mu_{w_2}) \in \mbR^{4}$. After a time discretization of these equations, the next state of the robot is distributed according to Gaussian distribution $\mcN(\bmu_{S},\bSig_{S})$
\begin{align}
    \mcS_{t} \sim q_{\varphi}(\mcS_{t}|\mcS_{t-1},u_{t-1}).
\end{align} 

% We introduce an inductive bias towards the kinematic structure, the (discrete) latent state transition may be parameterized by a neural network
% \begin{align}
%     &\bF_{1} = \begin{bmatrix}
%         \mu_{s1} + \mu_{w_1}\Delta t\\ \mbW_1 \sigma (\mbV_1 (\mu_{s_2}, \mu_{w_1},\mu_{s_2}))
%     \end{bmatrix}\\
%     &\bF_{2}\! =\! \begin{bmatrix}
%         \mu_{s_2} + \mu_{w_2}\Delta t\\ \mbW_2 \sigma (\mbV_2 (\mu_{s_2}, \mu_{w_1},\mu_{s_2}u))
%     \end{bmatrix} 
% \end{align} where $\sigma$ is a nonlinear activation function.
% The action state transition model and the IMU state transition model both represent the same physical state of the robot. 

On the other hand, the inertial observations from the IMU $\balpha(t)$ and the time-of-flight observations $\bX_{t}$ are encoded into Gaussian variables $\bA_{t}$ and $\bZ_{t}$  may also be used to estimate the state of the robot $\hat{\mcS}_t$ by considering the state transition function 
\begin{align}
    \hat{\mcS}_{t} \sim p_{\varphi}(\hat{\mcS}_{t}|\hat{\mcS}_{t-1},\balpha_{t-1},\bX_{t-1}).
\end{align} We desire that the latent state transition model for the actions $p_{\varphi}(\mcS_{t+1}|\mcS_{t},u_{t})$ should match the latent state transition model given from the sensors $p_{\varphi}(\hat{\mcS}_{t}|\hat{\mcS}_{t-1},\balpha_{t-1},\bX_{t-1})$. 

For the agent to have a sense of desirable actions, we introduce a reward distribution in order to predict its actions from its environment predictor 
\begin{align}
    \hat{r}_{t} \sim p_{\varphi}(\hat{r}_{t} | \mcS_{t},u_{t})
\end{align} which we model as a parameterized Gaussian distribution. The rewards are estimated   by the crawler to encourage forward locomotion. 

We model the perception system of the soft robotic crawler as process that minimizes the variational free energy stated as an optimization problem.

\begin{problem}[Perception] Given the parameterized state transition models and measurement models, we seek parameters $\bphi$ such that the variational free energy is minimized:
    \begin{align} \label{perception_cost} 
            &\mbox{argmin}_{\bphi} \,\mcF_{t}(\bphi) \\ \notag 
            &= \mathbb{E}_{q_{\bphi}}[D_{KL}[ q_{\varphi}(\mcS_{t}|\mcS_{t-1},u_{t-1})\, || \, p_{\varphi}(\hat{\mcS}_{t}|\hat{\mcS}_{t-1},\balpha_{t-1},\bX_{t-1})  ]] \\ \notag
            &\quad -\mathbb{E}_{q_{\bphi}}[\log\,p(\hat{\balpha}_{t},\hat{\bX}_{t}|\mcS_{t})] -\mathbb{E}_{q_{\bphi}}[\log\,p(\hat{r}_{t}|\mcS_{t})].  
    \end{align}
\end{problem} The first term measures the discrepancy between the state transition based on observations of the accelerations with the state transition model based on the control inputs. The second term measures the accuracy of the state transition model with the measurement model, and the third term predicts the reward received from taking an action.

\paragraph{Surrogate Perception Problem} The crawler receives observations from the IMU accelerometers $\balpha_{t}$ and the time-of-flight sensors $\bX_{t}$ which are distributed according to an unknown distribution $p(\bO_t) = p(\balpha(t),\bX(t))$. The IMU and TOF sensor observations are encoded into Gaussian distributions using the reparameterization trick \cite{doersch2016tutorial}
\begin{align}
    &\bA_{t} = \bmu(\balpha_{t},\bphi) + \bSig(\balpha_{t},\bphi)\odot\bepsilon_{\alpha}\\
    &\bZ_{t} = \bmu(\bX_{t},\bphi) + \bSig(\bX_{t},\bphi)\odot\bepsilon_{X}.
\end{align} where $\bphi$ are the model parameters, $\odot$ is element-wise multiplication, and $\bepsilon_{\alpha},\bepsilon_{X} \sim \mcN(\bzero,\bI)$ are zero mean Gaussian variables. 

To capture the time correlation between the states and the actions and for training stability of the state transition model, following \cite{hafner2019learning}, we introduce a hidden state $\bh_{t}$ and a recurrent neural network $f_{\varphi}$ to encode the previous state $\mcS_{t-1}$ and control input $u_{t-1}$ given by 
\begin{align}
    \bh_{t} = f_{\varphi}(\bh_{t-1},\mcS_{t-1},u_{t-1})
\end{align} that encodes the previous states and previous actions.

Therefore, we make the modification to the state transition models by conditioning on the hidden state $\bh_t$
\begin{align}
    &\mcS_{t} \sim q_{\varphi}(\mcS_{t}|\bh_{t})\\
    &\hat{\mcS}_{t} \sim p_{\varphi}(\hat{\mcS}_{t}|\bh_{t},\bA_{t},\bZ_{t})
\end{align} where $\bA_{t}$ and $\bZ_{t}$ are the Gaussian encoded IMU and TOF measurements, respectively. We model both the distributions $q_{\varphi}$ and $p_{\varphi}$ as neural network parameterized Gaussian distributions. The agent seeks to reconstruct its sensor observations $\bO_t = \left(\balpha_{t},\bX_{t} \right)$ from its latent states given by the distribution
\begin{align}
    \hat{\bO}_{t} \sim p_{\varphi}(\hat{\bO}_{t}|\mcS_{t},\bh_{t}).
\end{align}

% TK include remark on discretization

% TK mean approximation 
% TK mapping between the means 
% The relationship between the means of the robot state variables $\bmu= (\mu_{s_1},\mu_{\nu_1},\mu_{s_2},\mu_{\nu_2})$ and the mean IMU state variables using \eqref{mean_mapping:eq1}-\eqref{mean_mapping:eq2}. However, a priori the mapping \eqref{mean_mapping:eq1}-\eqref{mean_mapping:eq2} is unknown, hence we approximate this relationship with a parametrized function 
% \begin{align}
%     \bmu = F_{\btheta}(\tl{\bmu}).
% \end{align} for some parameters $\btheta$.

% Therefore, using these definitions, we define the parameterized state transition model for the dynamics by 
% \begin{align}
%     p_{\psi}(\bX_{t}|\bX_{t-1},u_{t-1}) 
% \end{align} defined by \eqref{eom_paremeterized:eq1} and \eqref{eom_paremeterized:eq2}. We define the encoding model for the state transition model given the observations by
% \begin{align}
%      q_{\btheta}(\bX_{t}|\bX_{t-1},\ba_{t-1}) = p(F_{\btheta}(\tl{\bmu}_{t})| F_{\btheta}(\tl{\bmu}_{t-1}),\ba_{t-1}).
% \end{align} The measurement model of the observations $\bY_{t} = (a_{1},a_{2},X_{1},X_{2})$ 
% \begin{align}
%     p(\bY_{t}|\bX_{t}) \sim \mcN(\bh(\bX_t),\bSig_{h}).
% \end{align} 

The agent seeks to minimize its surprise with respect to the state conditional on the observations and the state conditional on the previous state and action. The surprise is minimized by minimizing the following variational free energy functional

% TK on tricks for stability
To encourage numerical stability of the KL divergence, following \cite{hafner2020mastering,hafner2023mastering} we make the following modifications
\begin{align} \notag
    &\mcL_{1}(\bphi) = \max(I,D_{KL}[ \text{sg}\left(q_{\varphi}(\mcS_{t}|\bh_{t})\right)\, || \, p_{\psi}(\hat{\mcS}_{t}|\bh_{t},\bA_{t},\bZ_{t})  )]\\  \notag
    &\mcL_{2}(\bphi) = \max(I,D_{KL}[ \text{sg}\left(q_{\varphi}(\mcS_{t}|\bh_{t})\right)\, || \, \text{sg}\left(p_{\phi}(\hat{\mcS}_{t}|\bh_{t},\bA_{t},\bZ_{t}) \right)) ] 
\end{align} where $\text{sg}$ is the stop gradient function and a bound on the $I =2 \,\text{nats}$ is introduced to prevent model collapse ensuring that the latent dynamics has enough information about the inputs to learn nontrivial state transition dynamics. For compactness we define 
\begin{align}
    & \mcL_{3}(\bphi) = -\log\,p(\hat{\balpha}_{t},\hat{\bX}_{t}|\bh_{t},\mcS_{t}) -\log\,p(\hat{r}_{t}|\bh_{t},\mcS_{t}),
\end{align} Then, by introducing parameters $\beta_1, \beta_2$ and $\beta_3$ the modification of the perception cost function \eqref{perception_cost} is given by 
\begin{align} \label{surrogate_cost_perception}
    \tl{\mcF}(\bphi) = \mathbb{E}_{q_{\bphi}}[\beta_1 \mcL_{1}(\bphi) + \beta_1 \mcL_{2}(\bphi) + \beta_2 \mcL_{3}(\bphi)]
\end{align}

\subsection{Actor/Critic Learning}
The locomotion gait for a crawler is represented as a periodic control input $u_t$ given by 
\begin{align}\label{control_inputs}
    u_{t} = A_{0} + \sum^{N}_{k=1}A_{k}\sin(k\om t) + B_{k}\cos(k\om t)
\end{align} where $A_{k}$, $B_{k}$ and $\om$ are gait parameters to be learned by an actor-critic method. Let the gait parameters  $\ba = \left(A_{0},\dots,A_{N}, B_{1},\dots,B_{N},\om \right)$ be actions for the actor policy $\pi_{\theta}(\ba_{t}|\mcS_{t},\bh_{t})$.

For actor critic training we use the Dreamer algorithm \cite{hafner2019dream,hafner2020mastering,hafner2023mastering}. Given the learned latent state transition model $\mcS_{t}\sim p(\mcS_{t}|\bh_{t})$, the crawler uses the model to predict latent state trajectories over a prediction horizon and adapt its actions to maximize returns.  The actor aims to maximize the returns $R_{t} = \sum^{\infty}_{\tau=0}\gamma^{\tau}\,r^{t+\tau}$ with discount factor $\gamma$. Let $v_{\psi}(R_{t}|\mcS_{t},\bh_{t})$ be the critic (value) function which estimates the returns over the prediction horizon $H$. From replayed observations $(\bA_{t},\bZ_{t})$, the state transition model and actor generate a trajectory of learned latent states $\mcS_{1:H}$, actions $\ba_{1:H}$ and rewards $r_{1:H}$, the critic estimates a distribution of rewards $v_{\psi}(R_{t}|\mcS_{t},\bh_{t})$. Using predicted values $v_{t} := \mbE[v_{\psi}(\cdot |\mcS_{t}, \bh_{t})]$, the critic uses $\lambda$-bootstrapped returns 
\begin{align}
    &R_{t}^{\lambda} = r_{t} + \gamma\left((1-\lambda)v_{t} + \lambda R^{\lambda}_{t+1} \right), \quad t < H \\ 
    & R_{H}^{\lambda} = v_{H}
\end{align} the critic estimates the returns outside the prediction horizon $H$. The critic is trained to maximize the likelihood loss
\begin{align}\label{critic_loss}
    \mcL_{c}(\bpsi) = -\sum^{H}_{t=1}\log p_{\bpsi}(R_{t}^{\lambda}| \mcS_{t},\bh_{t})
\end{align}

\begin{figure}[H]
    \centering
    \includegraphics[width=0.90\linewidth]{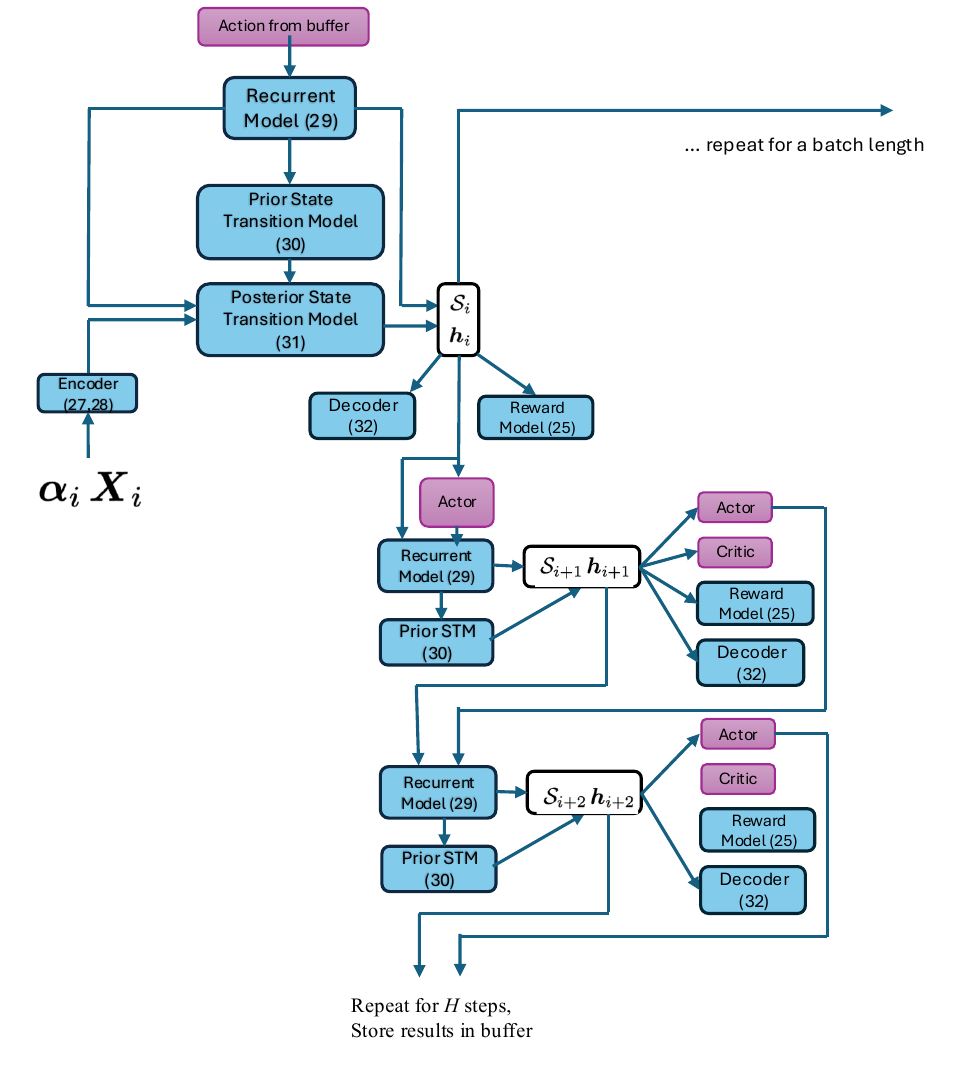}
    \caption{Summary of the training loop of the latent space state transition, actor and critic models.}
    \label{fig:placeholder}
\end{figure}

The actor policy $\pi_{\theta}(\ba_{t}|\mcS_{t},\bh_{t})$ is trained using Reinforced Gradients \cite{williams1992simple} over a batch of predicted latent states, observations, actions and rewards with surrogate loss function 
\begin{align}\label{actor_cost}
    \mcL_{a}(\btheta) = -\sum^{H}_{t=0}\mcA_{t} \log \pi_{\theta}(\ba_t|\mcS_{t},\bh_{t}) + \eta \bH[\pi_{\theta}(\ba_{t}|\mcS_{t},\bh_{t})]
\end{align} where $\bH[\pi_{\theta}]$ is the entropy of the policy which encourages exploration, $\eta = 10^{-4}$ is the entropy scale. The advantages  $\mcA_{t}$ are given by  
\begin{align}
    \mcA_{t} = \text{sg}\left(\frac{R_{t}^{\lambda} - v_{\psi}(\mcS_{t},\bh_{t})}{\max(1,S)} \right) 
\end{align} which are normalized by an exponential moving average of the difference between the 95$^{th}$ and 5$^{th}$ percentile over a batch of $R^\lambda_t$ returns
\begin{align}
    S = \text{EMA}(\text{Per}(R^{\lambda}_{t},95) - \text{Per}(R^{\lambda}_{t},5),\delta)
\end{align} using a decay factor of $\delta =0.99$.

\paragraph{Training Actor/Critic}  At each gradient step, we first we run the actor policy $\pi_{\theta}(\ba_{t}|\mcS_{t},\bh_{t})$ on the real environment to collect data. We store the observations, actions, and rewards $(\bO_{t},\ba_{t},r_{t},\bO_{t+1})$ in a replay buffer. Second, we sample sequences from the replay buffer to train the state-transition model, we encode the observations $\bO_{t} = (\balpha_{t},\bX_{t})$ into a Gaussian distribution $(\bA_{t},\bZ_{t})\sim\mcN((\bmu_{\alpha},\bmu_{X}),(\bSig_{\alpha},\bSig_{X}))$. Then using replayed trajectories, the state transition model $q_{\varphi}(\mcS_{t}|\bh_{t})$ for the latent states $\mcS_{t}$ along with its recurrent network $f_{\varphi}$ is trained by minimizing the surrogate cost function \eqref{surrogate_cost_perception} for the variational free energy. The third step uses the learned latent states $(\mcS_{t},\bh_{t})$ to forecast a trajectory over a prediction horizon of $H$ steps. The actor $\pi_{\theta}$ and the state transition model predicts the next states, and rewards producing a forecast trajectory $\mcT_{t:H} = (\mcS_{1:H}, \bh_{1:H},\ba_{1:H}, r_{1:H})$. 
Using the $H$-step trajectory, we compute the $\lambda$-returns and minimize the critic's predicted returns \eqref{critic_loss}. The actor is then updated to maximize the $\lambda$-returns using the loss function \eqref{actor_cost} using the predicted rollout $\mcT_{1:H}$. The three steps: 1) data collection, 2) state-transition training and 3) actor/critic training are repeated for a fixed number of gradient steps.

\section{Simulation Case Studies}\label{Simulation_studies}
We examine the latent-space reinforcement learning on the crawler system described in Section \ref{Crawler_system}. Following previous crawler models in the literature \cite{joey2017earthworm}, we choose the following physical parameters for the mass $m_1 = m_2 = 0.2\, \text{kg}$, the stiffness $k= 50 \, \text{N/m}$, damping $d = 0.1\, \text{N}\cdot \text{s/m}$, $F_{\sigma,1}=F_{\sigma,2} = 1$, and control weight $B_{u} = 1.0$. For the simulated IMU sensors, $\balpha(t)$, we assume a bias $b_a(t) = 0.05$ and zero mean Gaussian noise $\sigma_{\text{imu}} = 0.2$. The time-of-flight sensor, $\bX(t)$, has Gaussian with standard deviation $\sigma_{\text{tof}} = 0.2$. We choose gait parameters $\ba = \left(A_{0},A_{1},A_{2},B_{1},B_{2},\omega \right)$ to be learned by the actor network giving control inputs to the crawler given by \eqref{control_inputs}. The dynamics and sensors are simulated with a discrete step of $\Delta t = 0.01$ seconds. 

\begin{figure}[H]
    \centering
    \includegraphics[width=1.0\linewidth]{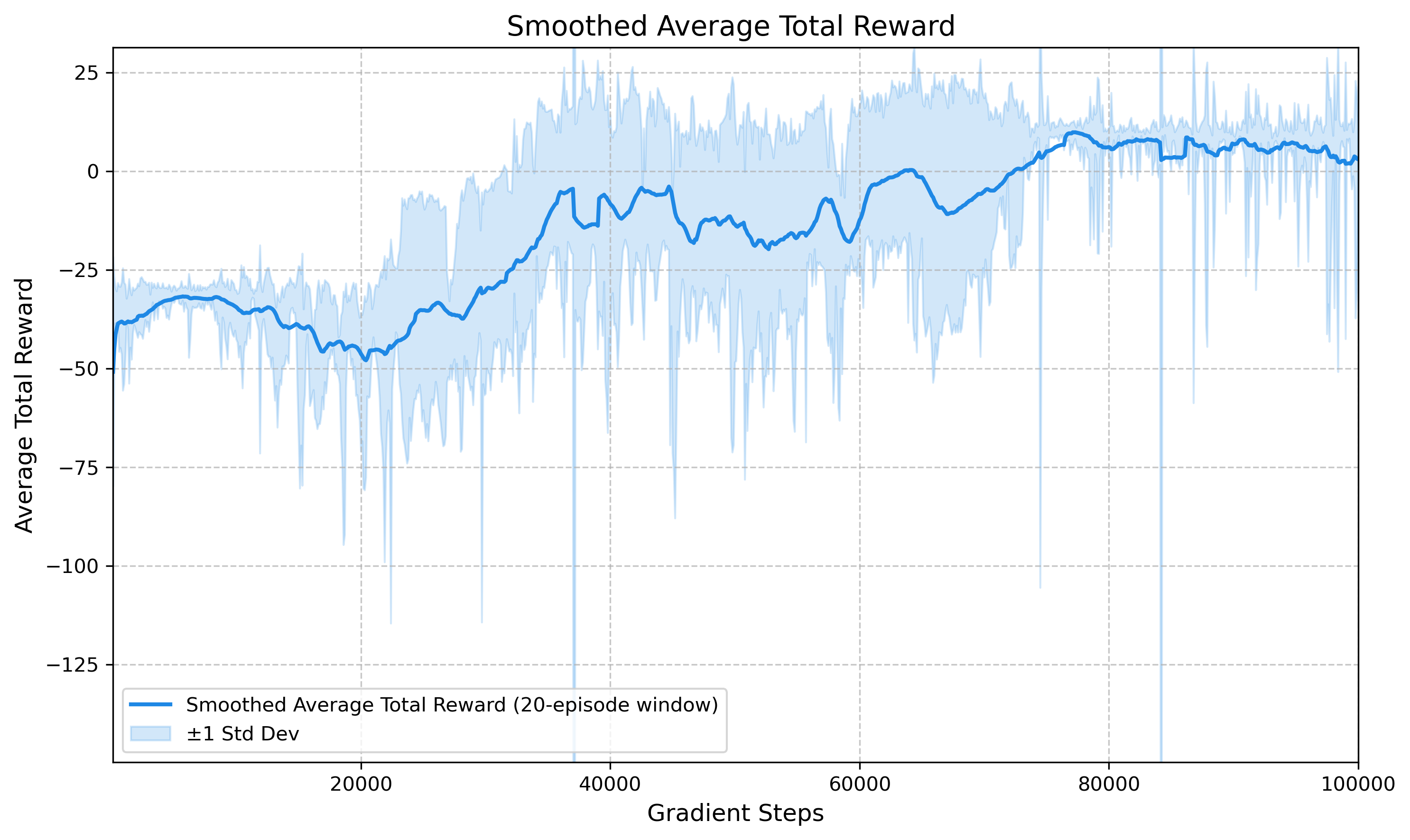}
    \caption{Average total rewards of 5 training runs with a moving average over 20 episodes over 100,000 gradient steps (solid blue). We include the variance over the training runs plotted in light blue.}
    \label{fig:metrics}
\end{figure}

%  The reward $r(\bs_t,\boldsymbol{w}_t,u_t)$ is for the locomotion of the robot described by the cost function $\mathcal{J}_T$
% \begin{align}
%     \mathcal{J}_T = \frac{1}{T}\left(w_1(T) - \alpha\int^{T}_{0} u(t)^2\, dt - \beta \int^T_{0}s_2(t)^2dt \right).
% \end{align} Over a time interval $[0,T]$, the agent seeks to maximize its forward locomotion $\nu_1(T)$ while minimizing the control effort $u(t)$ and the strain $s_2(t)$.

\begin{figure}[H]
    \centering
    \includegraphics[width=0.75\linewidth]{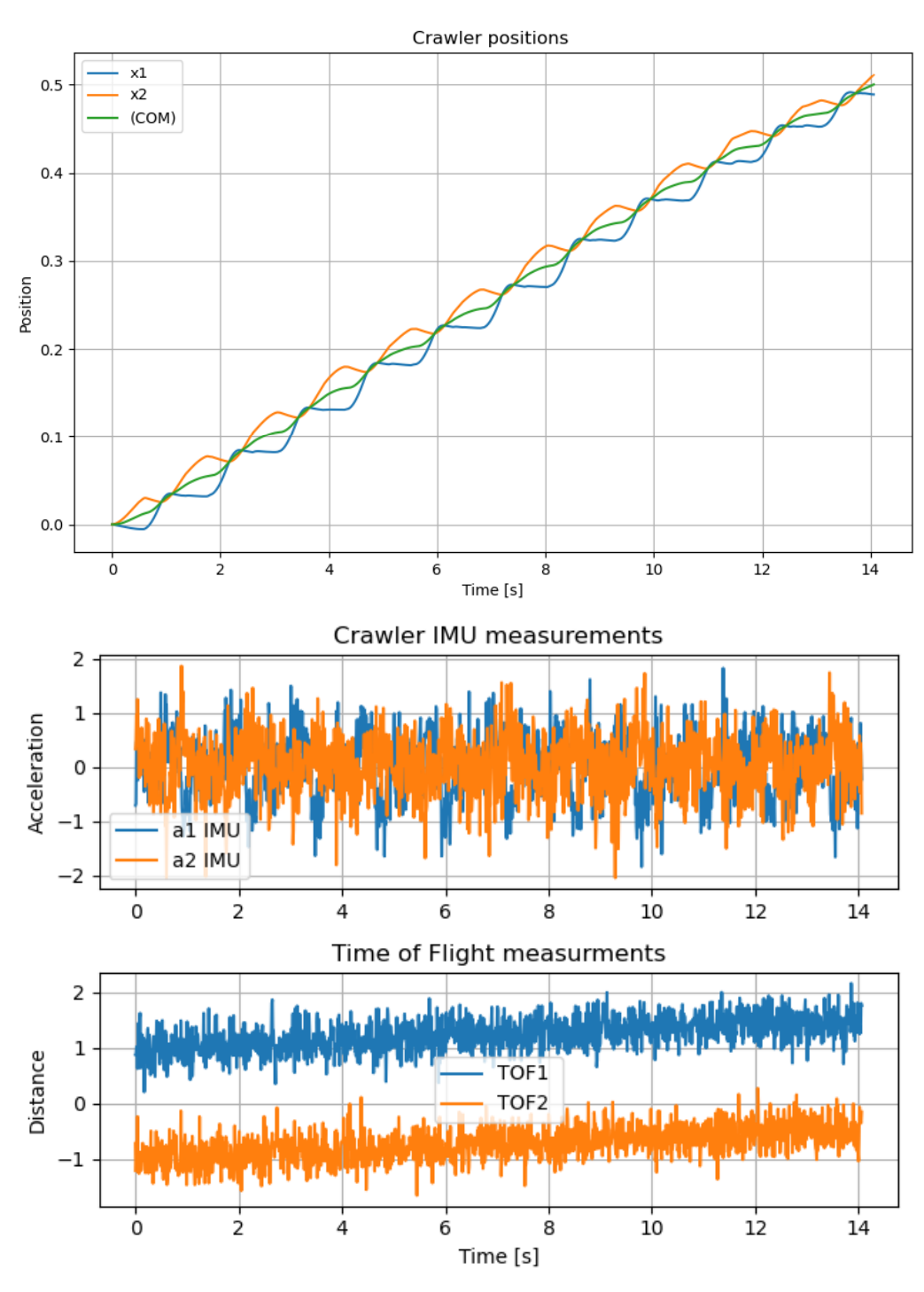}
    \caption{\textbf{Upper:} Location of the crawler under the locomotion policy. The head of the crawler is plotted in orange, the tail is plotted in blue, and the location of the centre of mass in green. \textbf{Lower:} Inertial measurement unit and time-of-flight observations}
    \label{fig:locomotion_policy_crawler}
\end{figure}

The goal of the crawler is to find a gait such that it moves to a specified target point as fast as possible.  The crawler is given the following rewards for locomotion to encourage learning a gait for forward locomotion 
\begin{align}
    r_1 = k_{1}\left(s_{1}(t) - s_1(t-\Delta t) \right) + k_{2} \max(0,w_{1}(t))
\end{align} with weights $k_1=0.25$, $k_2 = 0.5$. Furthermore, the crawler is given penalties for control effort,
\begin{align}
    r_2 &= -k_{3}|u(t)|^2 - k_{4}|u(t) - u(t-\Delta t)|^{2}\\ \notag
    & \quad -k_{5}|u(t)- u(t-2\Delta t)|^{2} 
\end{align} which takes into account the first and second order control rates, with weights $k_3 =k_4 = k_5 = 0.01$. We place a penalty on self-intersections 
\begin{align}
    r_4 = -k_{6}|s_2(t)|, \quad \text{if}\, s_2 <\eps
\end{align} for some positive threshold $\eps = 0.01$ and weight $k_6 = 0.01$. Lastly, we give a goal bonus when the crawler reaches the target of 50 cm forward within a 200 second episode, and a penalty if it fails 
\begin{align}
    r_{5} = \left \{\begin{aligned}&+2, \quad \text{if success}\\
    &-1, \quad \text{if failure.}\end{aligned} \right.
\end{align} We train the model with a prediction horizon of $H=15 \Delta t$ for 1000 iterations that divide 100,000 gradient steps and 100 replay steps with a batch size of $32$ and batch length $64$. The recurrent network has output state dimension of $512$, and have the following hyperparameters for the neural networks

\begin{table}[H]
    \centering
    \begin{tabular}{|c|c|c|c|}
        \hline
        Model & Layer Size & \# Layers & Learning Rate \\ \hline
        State Transition (Action) & 100 & 2 & 0.0001 \\ \hline
        State Transition (Obs) & 100 & 1 & - \\ \hline
        Recurrent & 100 & 1 & - \\ \hline
        Obs. Encoder & 100 & 2 & - \\ \hline
        Obs. Decoder & 100 & 2 & - \\ \hline
        Reward Decoder & 200 & 2 & - \\ \hline
        Actor & 200 & 2 & 0.00005 \\ \hline
        Critic & 200 & 2 & 0.00008 \\ \hline
    \end{tabular}
    \caption{List of hyperparameters}
    \label{tab:placeholder}
\end{table}

At each iteration, the buffer samples collected observations from the IMU and TOF sensors, actions and rewards $(\balpha_{1:T},\bX_{1:T},r_{1:T},\balpha_{1:T+1},\bX_{1:T+1},\ba_{1:T})$, to which the state transition model is trained, and then the actor/critic model is trained on the state-transition model. We observe in Figure \ref{fig:locomotion_policy_crawler}, that the learned policy is able to generate an effective gait for the crawler that reaches its goal of forward locomotion in approximately $14$ seconds. In Figure \ref{fig:metrics}, we compute the average reward over 5 training runs, and place a sliding window of 20 episodes. We also observe in Figure \ref{fig:metrics}  that crawler is able to learn a latent space model from its sensors that finds a gait that maximizes its returns.

\section{Conclusion}\label{Conclusion}
This paper studied  latent space approach to model-based reinforcement learning for soft crawler locomotion using noisy sensor observations. We developed a Gaussian state transition model for the crawler based on minimizing a surrogate for the variational free energy. The state transition model is used to train  actor/critic networks to find a gait that moves the crawler to a target as fast as possible. In future work, we consider other physics informed latent space representations of the soft robot state transition model and explore architecture-optimizer co-design to improve convergence. 
We believe this line of research can advance embodied intelligence, enabling soft robots to plan and adapt gaits with greater autonomy and robustness.

% \clearpage

% \begin{align}
%     \mcS_{i}\\
%     \bh_{i}
% \end{align}

% \begin{align}
%     \mcS_{i+1}\, \bh_{i+1}
% \end{align}

% \begin{align}
%     \mcS_{i+2}\, \bh_{i+2}
% \end{align}

% \begin{align}
%     \balpha_{i} \, \bX_{i}
% \end{align}

\printbibliography

@book{marsden2013introduction,
  title={Introduction to mechanics and symmetry: a basic exposition of classical mechanical systems},
  author={Marsden, Jerrold E and Ratiu, Tudor S},
  volume={17},
  year={2013},
  publisher={Springer Science \& Business Media}
}

@article{williams1992simple,
  title={Simple statistical gradient-following algorithms for connectionist reinforcement learning},
  author={Williams, Ronald J},
  journal={Machine learning},
  volume={8},
  number={3},
  pages={229--256},
  year={1992},
  publisher={Springer}
}

@article{shepherd2011multigait,
  title={Multigait soft robot},
  author={Shepherd, Robert F and Ilievski, Filip and Choi, Wonjae and Morin, Stephen A and Stokes, Adam A and Mazzeo, Aaron D and Chen, Xin and Wang, Michael and Whitesides, George M},
  journal={Proceedings of the national academy of sciences},
  volume={108},
  number={51},
  pages={20400--20403},
  year={2011},
  publisher={National Academy of Sciences}
}

@article{sun2021soft,
  title={Soft mobile robots: A review of soft robotic locomotion modes},
  author={Sun, Yinan and Abudula, Aihaitijiang and Yang, Hao and Chiang, Shou-Shan and Wan, Zhenyu and Ozel, Selim and Hall, Robin and Skorina, Erik and Luo, Ming and Onal, Cagdas D},
  journal={Current Robotics Reports},
  volume={2},
  number={4},
  pages={371--397},
  year={2021},
  publisher={Springer}
}

@inproceedings{jilani2024solving,
  title={Solving dynamic cosserat rods with frictional contact using the shooting method and implicit surfaces},
  author={Jilani, Radhouane and Villard, Pierre-Fr{\'e}d{\'e}ric and Kerrien, Erwan},
  booktitle={2024 IEEE/RSJ International Conference on Intelligent Robots and Systems (IROS)},
  pages={10483--10488},
  year={2024}}

@article{zhang2024adaptive,
  title={An adaptive lumped-mass dynamic model and its control application for continuum robots},
  author={Zhang, Xu and Yang, Chenghao and Song, Zhibin and Khanesar, Mojtaba A and Branson, David T and Dai, Jian S and Kang, Rongjie},
  journal={Mechanism and Machine Theory},
  volume={201},
  pages={105736},
  year={2024},
  publisher={Elsevier}
}

@article{zhang2021inchworm,
  title={Inchworm inspired multimodal soft robots with crawling, climbing, and transitioning locomotion},
  author={Zhang, Yifan and Yang, Dezhi and Yan, Peinan and Zhou, Peiwei and Zou, Jiang and Gu, Guoying},
  journal={IEEE Transactions on Robotics},
  volume={38},
  number={3},
  pages={1806--1819},
  year={2021},
  publisher={IEEE}
}

@article{ferrentino2023finite,
  title={Finite element analysis-based soft robotic modeling: Simulating a soft actuator in sofa},
  author={Ferrentino, Pasquale and Roels, Ellen and Brancart, Joost and Terryn, Seppe and Van Assche, Guy and Vanderborght, Bram},
  journal={IEEE robotics \& automation magazine},
  volume={31},
  number={3},
  pages={97--105},
  year={2023},
  publisher={IEEE}
}

@article{zhang2019design,
  title={Design and modeling of a parallel-pipe-crawling pneumatic soft robot},
  author={Zhang, Zhiyuan and Wang, Xueqian and Wang, Songtao and Meng, Deshan and Liang, Bin},
  journal={IEEE access},
  volume={7},
  pages={134301--134317},
  year={2019},
  publisher={IEEE}
}

@article{lin2023single,
  title={Single-actuator soft robot for in-pipe crawling},
  author={Lin, Ying and Xu, Yi-Xian and Juang, Jia-Yang},
  journal={Soft Robotics},
  volume={10},
  number={1},
  pages={174--186},
  year={2023},
  publisher={Mary Ann Liebert, Inc., publishers 140 Huguenot Street, 3rd Floor New~…}
}

@article{pan2025bioinspired,
  title={Bioinspired Mechanisms and Actuation of Soft Robotic Crawlers},
  author={Pan, Min and Liu, Miaomiao and Lei, Jiayi and Wang, Yunyi and Linghu, Changhong and Bowen, Chris and Hsia, K Jimmy},
  journal={Advanced Science},
  volume={12},
  number={16},
  pages={2416764},
  year={2025},
  publisher={Wiley Online Library}
}

@inproceedings{joey2017earthworm,
  title={An earthworm-inspired soft crawling robot controlled by friction},
  author={Joey, Z Ge and Calder{\'o}n, Ariel A and P{\'e}rez-Arancibia, N{\'e}stor O},
  booktitle={2017 IEEE international conference on robotics and biomimetics (ROBIO)},
  pages={834--841},
  year={2017}}

@article{shen2024optimal,
  title={Optimal gait design for nonlinear soft robotic crawlers},
  author={Shen, Yenan and Leonard, Naomi Ehrich and Bamieh, Bassam and Arbelaiz, Juncal},
  journal={IEEE Control Systems Letters},
  year={2024},
  publisher={IEEE}
}

@article{chen2001locomotive,
  title={Locomotive gait generation for inchworm-like robots using finite state approach},
  author={Chen, I-Ming and Yeo, Song Huat and Gao, Yan},
  journal={Robotica},
  volume={19},
  number={5},
  pages={535--542},
  year={2001},
  publisher={Cambridge University Press}
}

@inproceedings{even2023locomotion,
  title={Locomotion and obstacle avoidance of a worm-like soft robot},
  author={Even, Sean and Ozkan-Aydin, Yasemin},
  booktitle={2023 IEEE/RSJ international conference on intelligent robots and systems (IROS)},
  pages={6491--6496},
  year={2023}}

@article{riddle20253d,
  title={A 3D Model Predicts Behavior of a Soft Bodied Worm Robot Performing Peristaltic Locomotion},
  author={Riddle, Shane and Jackson, Clayton B and Daltorio, Kathryn A and Quinn, Roger D},
  journal={Bioinspiration \& Biomimetics},
  year={2025}
}

@inproceedings{gillespie2018learning,
  title={Learning nonlinear dynamic models of soft robots for model predictive control with neural networks},
  author={Gillespie, Morgan T and Best, Charles M and Townsend, Eric C and Wingate, David and Killpack, Marc D},
  booktitle={2018 IEEE International Conference on Soft Robotics (RoboSoft)},
  pages={39--45},
  year={2018}}

@inproceedings{asawalertsak2023small,
  title={A small soft-bodied crawling robot with electromagnetic legs and neural control for locomotion on various metal terrains},
  author={Asawalertsak, Naris and Nantareekurn, Worameth and Manoonpong, Poramate},
  booktitle={2023 IEEE International Conference on Soft Robotics (RoboSoft)},
  pages={1--8},
  year={2023}}

@article{doersch2016tutorial,
  title={Tutorial on variational autoencoders},
  author={Doersch, Carl},
  journal={arXiv preprint arXiv:1606.05908},
  year={2016}
}

@article{chua2018deep,
  title={Deep reinforcement learning in a handful of trials using probabilistic dynamics models},
  author={Chua, Kurtland and Calandra, Roberto and McAllister, Rowan and Levine, Sergey},
  journal={Advances in neural information processing systems},
  volume={31},
  year={2018}
}

@article{asawalertsak2023frictional,
  title={Frictional anisotropic locomotion and adaptive neural control for a soft crawling robot},
  author={Asawalertsak, Naris and Heims, Franziska and Kovalev, Alexander and Gorb, Stanislav N and J{\o}rgensen, Jonas and Manoonpong, Poramate},
  journal={Soft Robotics},
  volume={10},
  number={3},
  pages={545--555},
  year={2023},
  publisher={Mary Ann Liebert, Inc., publishers 140 Huguenot Street, 3rd Floor New~…}
}

@article{gusty2025optimal,
  title={Optimal Control of Soft-Robotic Crawlers Subject to Nonlinear Friction: A Perturbation Analysis Approach},
  author={Gusty, Andrew and Scarborough, Cody and Arbelaiz, Juncal and Jensen, Emily},
  journal={IEEE Control Systems Letters},
  year={2025},
  publisher={IEEE}
}

@inproceedings{chang2021shape,
  title={Shape-centric modeling for soft robot inchworm locomotion},
  author={Chang, Alexander H and Freeman, Caitlin and Mahendran, Arun Niddish and Vikas, Vishesh and Vela, Patricio A},
  booktitle={2021 IEEE/RSJ International Conference on Intelligent Robots and Systems (IROS)},
  pages={645--652},
  year={2021}}

@inproceedings{hafner2019learning,
  title={Learning latent dynamics for planning from pixels},
  author={Hafner, Danijar and Lillicrap, Timothy and Fischer, Ian and Villegas, Ruben and Ha, David and Lee, Honglak and Davidson, James},
  booktitle={International conference on machine learning},
  pages={2555--2565},
  year={2019},
  organization={PMLR}
}

@inproceedings{wu2023daydreamer,
  title={Daydreamer: World models for physical robot learning},
  author={Wu, Philipp and Escontrela, Alejandro and Hafner, Danijar and Abbeel, Pieter and Goldberg, Ken},
  booktitle={Conference on robot learning},
  pages={2226--2240},
  year={2023},
  organization={PMLR}
}

@article{li2025robotic,
  title={Robotic world model: A neural network simulator for robust policy optimization in robotics},
  author={Li, Chenhao and Krause, Andreas and Hutter, Marco},
  journal={arXiv preprint arXiv:2501.10100},
  year={2025}
}

@article{matsuo2022deep,
  title={Deep learning, reinforcement learning, and world models},
  author={Matsuo, Yutaka and LeCun, Yann and Sahani, Maneesh and Precup, Doina and Silver, David and Sugiyama, Masashi and Uchibe, Eiji and Morimoto, Jun},
  journal={Neural Networks},
  volume={152},
  pages={267--275},
  year={2022},
  publisher={Elsevier}
}

@article{hafner2023mastering,
  title={Mastering diverse domains through world models},
  author={Hafner, Danijar and Pasukonis, Jurgis and Ba, Jimmy and Lillicrap, Timothy},
  journal={arXiv preprint arXiv:2301.04104},
  year={2023}
}

@article{hafner2020mastering,
  title={Mastering atari with discrete world models},
  author={Hafner, Danijar and Lillicrap, Timothy and Norouzi, Mohammad and Ba, Jimmy},
  journal={arXiv preprint arXiv:2010.02193},
  year={2020}
}

@article{hafner2019dream,
  title={Dream to control: Learning behaviors by latent imagination},
  author={Hafner, Danijar and Lillicrap, Timothy and Ba, Jimmy and Norouzi, Mohammad},
  journal={arXiv preprint arXiv:1912.01603},
  year={2019}
}

@inproceedings{deisenroth2011pilco,
  title={PILCO: A model-based and data-efficient approach to policy search},
  author={Deisenroth, Marc and Rasmussen, Carl E},
  booktitle={Proceedings of the 28th International Conference on machine learning (ICML-11)},
  pages={465--472},
  year={2011}
}

@article{garcia1989model,
  title={Model predictive control: Theory and practice—A survey},
  author={Garcia, Carlos E and Prett, David M and Morari, Manfred},
  journal={Automatica},
  volume={25},
  number={3},
  pages={335--348},
  year={1989},
  publisher={Elsevier}
}

@article{gamus2020understanding,
  title={Understanding inchworm crawling for soft-robotics},
  author={Gamus, Benny and Salem, Lior and Gat, Amir D and Or, Yizhar},
  journal={IEEE Robotics and Automation Letters},
  volume={5},
  number={2},
  pages={1397--1404},
  year={2020},
  publisher={IEEE}
}

@book{marsden1992lectures,
  title={Lectures on mechanics},
  author={Marsden, Jerrold E},
  volume={174},
  year={1992},
  publisher={Cambridge University Press}
}

@article{das2023earthworm,
  title={An earthworm-like modular soft robot for locomotion in multi-terrain environments},
  author={Das, Riddhi and Babu, Saravana Prashanth Murali and Visentin, Francesco and Palagi, Stefano and Mazzolai, Barbara},
  journal={Scientific Reports},
  volume={13},
  number={1},
  pages={1571},
  year={2023},
  publisher={Nature Publishing Group UK London}
}

@article{marsden2006hamiltonian,
  title={Hamiltonian reduction by stages},
  author={Marsden, Jerrold E and Misiolek, Gerard and Ortega, Juan-Pablo and Perlmutter, Matthew and Ratiu, Tudor S},
  journal={Lecture Notes in Mathematics},
  year={2006}
}
% \bibliographystyle{IEEEtran}
%  \bibliography{bibliography}

\end{document}